# Optimising Rule-Based Classification in Temporal Data

Polla Fattah[1,2*], Uwe Aickelin[2] and Christian Wagner[2]

[1]College of Engineering, Salahaddin University-Erbil, Erbil, Kurdistan region, Iraq
[2]School of Computer Science, University of Nottingham, Nottingham, UK

## ABSTRACT

This study optimises manually derived rule-based expert system classification of objects according to changes in their properties over time. One of the key challenges that this study tries to address is how to classify objects that exhibit changes in their behaviour over time, for example how to classify companies' share price stability over a period of time or how to classify students' preferences for subjects while they are progressing through school. A specific case the paper considers is the strategy of players in public goods games (as common in economics) across multiple consecutive games. Initial classification starts from expert definitions specifying class allocation for players based on aggregated attributes of the temporal data. Based on these initial classifications, the optimisation process tries to find an improved classifier which produces the best possible compact classes of objects (players) for every time point in the temporal data. The compactness of the classes is measured by a cost function based on internal cluster indices like the Dunn Index, distance measures like Euclidean distance or statistically derived measures like standard deviation. The paper discusses the approach in the context of incorporating changing player strategies in the aforementioned public good games, where common classification approaches so far do not consider such changes in behaviour resulting from learning or in-game experience. By using the proposed process for classifying temporal data and the actual players' contribution during the games, we aim to produce a more refined classification which in turn may inform the interpretation of public goods game data.

## 1. INTRODUCTION

Due to the rapid accumulation of temporal data in various fields like economy, medicine and education (Deshpande & Karypis 2002; Wei & Keogh 2006; Lane & Brodley 1999; Sebastiani 2002), the need for understanding them and grouping them through their patterns, similarities and differences was raised some decades ago, and several data mining methods like classification and clustering for temporal data have been presented to find useful information beyond static snapshots (Sonnenburg et al. 2005). However, these methods have two problems: first, they ignore the behaviour of individual objects at each time point, focussing instead on the time series as a whole (Keogh & Pazzani 2000); and second, most of these methods have a common problem with non-temporal (traditional) data classifiers, involving human interpretability for the generated rules (Xing et al. 2011; Ishibuchi & Nojima 2007; Letham et al. 2010).

The targeted data for this study is temporal data representing objects' behaviour through time. Each object might have one or multiple temporal attributes associated with different aspects of their behaviour. Examples for this data include companies' share price behaviour in the stock market (e.g. risky or stable), students' subject tendencies in different stages of school (science, literacy, art or mathematics), or athletic performance in a series of games. However, the case study in this paper is classifying players according to their contribution in the public goods game experiment, conducted by Fischbacher et al. (Fischbacher et al. 2012). Public goods game is an experimental simulation for real-life public goods which tries to find different human contribution behaviours (Dufwenberg et al. 2011).

In this paper we propose a new method to classify temporal data by optimising human-generated rules for existing classes. The proposed classifier consists of two main stages, first rule generation and second rule optimisation (see Figure 1).

In the first stage, numerous classifier rules are generated using expert definitions and opinions about existing classes regarding specific objects we are interested in classifying. Each class might be represented by multiple attributes reflecting class definition as presented by the experts. These attributes are mostly aggregated and non-temporal attributes derived from temporal attributes of objects such as human expert understandings, and classes are defined using aggregation and generalisation. Depending on definitions and various experts' opinions, each attribute can be limited in a range of values for each class. These ranges for multiple attributes are the source of creating classifiers with different boundaries for each class.

The goal of the optimisation stage is select a classifier which can create the most compacted objects for each class in every time point. The compactness of classes can be measured using different criteria like standard deviation, Euclidian distance and internal clustering indices. Smaller distances between objects at every time point indicate better classier limits. This stage reduces the initial boundary ranges of values between classes into single crisp values. The resultant classes reflect both general human definition and understanding for the classes and specific group of objects' behaviour in individual time points. Such optimisation could be achieved using heuristic methods, but in this paper we used brute forcing to produce the best possible results.

In our case study (Fischbacher et al. 2012) economists used non-temporal data called contribution table to classify players into four groups: free riders, conditional contributors, tringle contributors and others (Fischbacher et al. 2001). A contribution table is a table filled by the players prior to the start of the game to represent their initial willingness to contribute in response to the other players in the group. This classification ignores players' actual amount of contribution and change of strategy during game rounds.

By using the proposed method for classification, we can include players' contributions and beliefs about other players' contributions in the actual games as temporal data to classify them. These temporal elements might provide new information to the classifier conveying players' strategy throughout the game that was unavailable for the previous classifier. For the first stage, we used a modified version of the original definitions for classes reflecting time element. The classifiers produced are optimised using brute force method to find the best classifier with a cost function depending on the standard deviation of the contributions at each time point.

New classes are compared with those of the economists by classifying players using them as training and testing labels for support vector machine (SVM) classifier. We provided it with different sets of attributes, some of which had not been seen by our classifier or that of the economists. According to the area under the curve AUC measures, the labels provided by the proposed classifier scored higher values than those of the economists.

## 2. RELATED WORKS

Classifying data using time dimension can be achieved using existing classifiers like SVM and ID3 (Revesz & Triplet 2011). In their work, they tried to take advantage of time dimension in their data by shifting data features for a specified number of time points (n time point). By shifting they mean changing up to n time points to new attributes, so that traditional classifiers like SVM and ID3 can look for these extra days as features to be considered in the classification. What we are looking for in our system is to classify objects with consideration of all time points and without having initial classes for training the classifier model. (Kadous 2002) used feature construction from sub-events to create features for the classifier, and they introduced TClass for temporal classification. However, these models need a training dataset to be labelled to start with, while we are trying to start from general definitions to create a classifier and produce more specified definitions for each class.

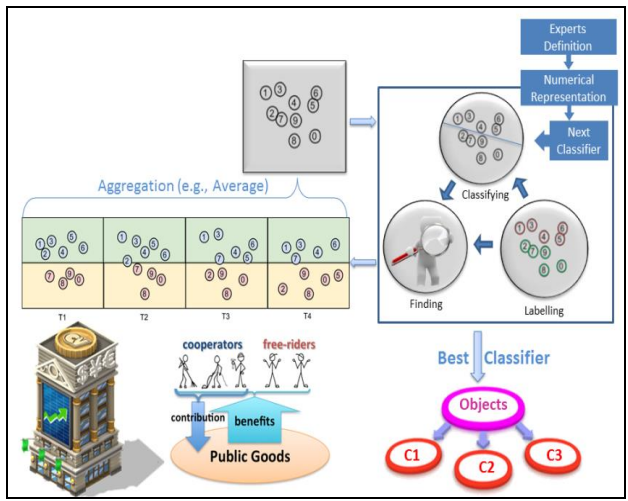

Figure 1. An illustration for items' attribute value change in temporal data and how our proposed system takes this change in consideration.

Another approach for classifying temporal data is to feed the whole time series elements as features to the classifiers like SVN and K nearest neighbour (KNN) and use distance measures to classify them. Some methods used traditional distance measures like Euclidean and Manhattan measures (Wei & Keogh 2006; Keogh & Kasetty 2003), while Keogh et al. (Keogh & Pazzani 2000) proposed Dynamic Time Warping (DTW), which measures the distance between two time series by elastically finding similarities between them instead of measuring each time point separately. These methods consider the whole time series as one entity and try to find distance between them using different measures, while in the proposed method we use dimensions between classified items in each time point separately to optimise manually created classifiers.

Considerable research has been undertaken to track behaviour changes in data streams using clustering (Spiliopoulou et al. 2006; Günnemann et al. 2011; Hawwash & Nasraoui 2012; Kalnis et al. 2005; Aggarwal 2005), but all of these methods focussed on the clusters in general (e.g. how clusters are changing, merging, moving and disappearing) or data concentration in the data streams, and they did not explore individual objects in a single slice of time period. We are trying to use both the general aspect, represented by time aggregated attributes of each time series, and separate time points for each individual, by calculating the distances of each class for every time point.

## 3. BACKGROUND

To case study data originated from an economics experiment for a public goods game (Fischbacher et al. 2012), and we will be comparing our results to their classes. The next section briefly describes the public goods game and economists' classification method for player types, and a short review for used techniques and algorithms like classification, clustering indices and area under the curve are presented in later sections.

### A. Public Goods Game

Public good is a resource or service that cannot be restricted from access by individuals; there is no need to compete for such resources and services as they are characterised by "non-rivalry and non-excludability" (Kaul et al. 1999). Examples of public good include public parks which anyone can attend, regardless of their contribution to taking care of it, or street lights and other civic amenities. Another example which is an exclusive category of the public good is a student apartment's dishwasher. In the last example, every student in the apartment can use the dishwasher, but it is exclusively used among students of the apartment.

So public goods game or public goods experiment is a simulation of the real situation of public good in a lab with limited conditions and focused purposes to conduct experiments that can be studied and measured (Dufwenberg et al. 2011), so that public goods game can be considered as a non-zero sum game. There are different set-ups for this experiment, all of which are similar in terms of being multiple player games. Each player starts with a certain amount of points which represent real money. Players have to make to decide what amount of points they will contribute in a virtual public good (a project). Each individual player's contribution can be between zero and the maximum amount that the player can contribute. The total amount of contribution is concealed from players until the end of the round. A round of the game will be finished after all players decide their amount of contribution, then each player will receive an equal share of the benefit from the project regardless of their amount of contribution. The benefit is in the form of points replaced by a percentage of real money later on.

Our study follows players' categorisation as defined by (Fischbacher et al. 2001), with whose results we will compare our own later. According to them, there are three main categories for public goods game players and one category is called 'others', which is for players who do not fit in any of their definitions. These categories are:

**Conditional Co-Operator**: these players how show more willingness to contribute when other players contribute more.

**Free Riders**: these players do not contribute to the project regardless of other players' contribution status.

**Triangle Contributors**: these players' contribution will rise to a point then start to decline afterward regarding other players' contribution.

**Others**: these players do not have a clear pattern in their contribution style.

*B. Classification and Performance Measures*

Rule-based classification is a set of if-else rules provided by human experts to identify existing items into a certain category by using items' attributes as condition variables to sort out items' labels (Negnevitsky 2002). In this paper, we use rule-based classification on the aggregated attributes to classify items before measuring its fitness for all time points by using cost function.

SVM is a popular classification method that creates hyper plains among different classes (Furey et al. 2000). In this paper we use SVM to compare between proposed systems' generated labels against economists' labels by training SVM model using a portion of data, then testing the trained model on the unseen part of the data by the classifier. For comparison purposes we created new attributes derived from existing ones not used by any classifiers (i.e. proposed classifier and economists' classification), so that SVM can classify items independently according to the previous attributes, which contributed in generating labels directly. For validating SVM results we use area under the curve to measure performance of SVM with both labels.

*C. Area under the Curve*

Area Under the Curve (AUC) of receiver operating characteristic (ROC) is a measure used by (Bradley 1997) to calculate the performance of machine learning algorithms such as classification. The ROC curve is a graph of true positive rate TPR and false positive rate FPR of the predicted classifier's result compared to the real class for each item, so that AUC is the area under that curve. Methods of calculating AUC vary according to the nature of application and available data. The multi-class AUCs are calculated using the following equation (Hand & Till 2001):

$$auc = \frac{2}{c(c-1)} \sum aucs$$

Where c is number of classes and aucs is a set of auc between any two classes.

*D. Internal Cluster Indices*

Internal cluster measure makes use of information provided by underlying data and suggested clusters such as compactness of data points inside one cluster and separation of clusters from each other to assess validity of clustering algorithm and provide a numeric measure of how well the clusters are grouping data examples of internal indices are Dunn Index, Silhouette Coefficient and Davies–Bouldin Index (Rendón et al. 2011).

## 4. METHODOLOGY

The proposed method for classification consists of two steps. The first step uses experts' knowledge in a specific area of expertise to create a set of multiple rule-based classifiers. These classifiers are similar in the rules, however they apply different values for these rules. The used values in this step can be derived by aggregating from the temporal data. The second step tries to find the best classifier among the set of classifiers by selecting a specific value for each rule from provided range. This task is a typical task of optimisation. Figure 2 shows all parts of the flowchart explained in the subsequent sections. For the purpose of illustration, we will solve a simplified classification problem using proposed system. Suppose we have data about students' grades across multiple classes. We have been asked to classify the studnets into bad, good and excellent according to their grades. In the next few sections we will try to achieve this while demonstrating each step of the system in more detail.

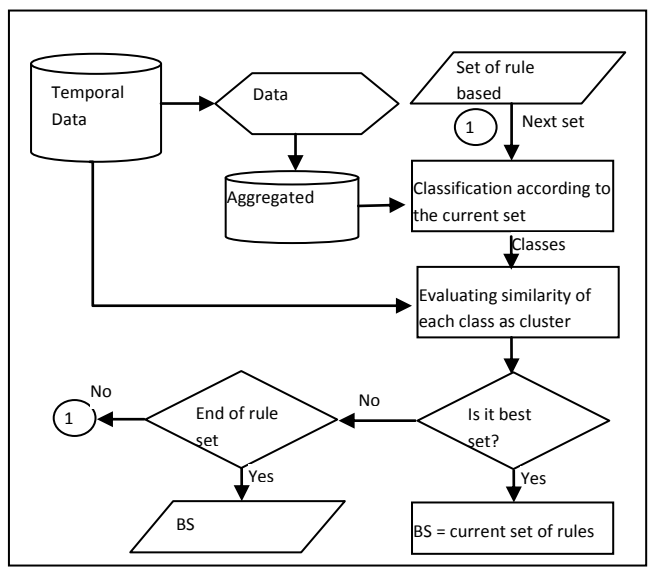

Figure 2. Flowchart explaining proposed system steps for optimising rule-based classification.

### A. Targeted Data for Classification

The targeted data for this classification is time series which is a sequence of a univariate or multivariate observations obtained at different time points (Lee et al. 2004; Yang & Shahabi 2004). The time series used for this classifier should have equal and discrete time points. New attributes for the items can be added later as an aggregated attribute derived from the obtained temporal data.

### B. Generating Rule-Based Classifiers

To generate rule-based algorithms for a dataset the number of classes and their labels will be determined by subject-specific experts. The knowledge of experts can be in the form of direct interaction with the system to determine how many classes should be used for the underlying data, suitable labels for each class and the rules used by the experts themselves to differentiate between them. Another way to determine classes and their boundaries is to use experts' general definitions for the items to be classified. The third way is a mix between the two previous methods, starting by the definition of the classes then asking exerts to fill the gaps and clarify ambiguities, or to ascertain their opinion about a specific dataset.

To have rules for the classes that are interpretable by humans and easy to understand, the aggregated features of the individual items should be used, derived from obtained time series. Examples of aggregation are minimum value, maximum value, mean, mode, median and standard deviation of time values of each object across all time points. Moreover, the rules should use mathematical comparison signs like $\{<, >$ or $=\}$ to separate classes from each other, comparing aggregated attributes limits for each class.

The aggregated attributes used in the definitions for the classes should have upper and lower limits for each class. These limits represent an expected range of values for the attributes to consider an item falls in that class. Moreover, the initial value of the upper limit for a class might overlap with the value of the lower limit of the next class. This range of values and their overlaps may originate from differences in experts' opinions or from slightly different definitions for each class. The next step focuses on reducing each range of these values to a single value that represents a split between two classes and eliminates overlaps.

In the students example, the recorded marks of subjects on which the student was examined in multiple years can be aggregated to have an average mark for each student separately, then the experts (teachers in this example) determine the student averages and assign them to particular classes, thus:

- If a student has an average mark of 58 or lower, then she can be considered as a bad student.
- If the student has an average mark between 55 to 67, she can be considered as a good student
- If a student has an average mark higher than 65, she can be considered as an excellent student.

It can be seen that there are overlaps in these values, which can be represented using comparison signs and placing upper and lower limits in a vector to represent the range of initially acceptable values:

- If average(mark) $\geq$ [65, 100] then: Excellent student
- If average(mark) $\leq$ [50, 58] then: Bad student
- Else: Good Student

Next in the optimization stage we will reduce these ranges to represent a single value.

*C. Optimizing Generated Rules*

This stage optimizes rules generated in the first stage using experts' opinions about number of classes and the range of values which divides each class from its neighbours. This step uses optimization to narrow down the provided range of values into a single value that divides classes' boundaries. This single value represents the best dividing point between these classes. The point is considered as the best dividing point when it produces the most compacted classes of items in every time point. Algorithm 1 represents this process, which is discussed in more detail in the following paragraphs.

The process of optimisation can be accomplished by iterating through all possibilities of the value ranges for the rule based classifiers to classify the underlying data using aggregated attributes then evaluating the results by calculating compactness of classes in every time point using the temporal data. This means the optimisation step can be divided into two sub-steps, classification and evaluation.

**Data**: Temporal data and aggregated attributes to represent classification rules.
**Data:** R= set of classification rules which includes discrete value rages.
**Data:** minCost = Inf.

**function** *calculateCost(C)*:
　　*costs* = vector
　　**for** *t in Periods* **do**
　　　　**for** *c in Classes* **do**
　　　　　　costs.append($CM(c^t)$ * count(c))
　　**return** *sum(costs)*

**for** *r in R* **do**:
　　c = classify(PG, r)
　　cost = calculateCost(c)
　　**if** *cost < minCost* **then**:
　　　　minCost = cost
　　　　bestClassifier = r
　　**return** bestClassifier

Algorithm 1 Process of optimizing provided rules using brute force

The classification step uses provided rules with a single value for each range of the values. If the value ranges are continuous they should be discretised into acceptable discrete values. Selecting the acceptable discretisation intervals is subject of specific area and underlying data which can be decided with consultancy of area specific experts. By iterating though all values the classifier tries values to classify underlying data labels items accordingly and sends them to the next step to be evaluated.

The evaluation step uses item labels provided by the classifier of the previous step and uses temporal attributes to evaluate compactness of the classes in each time point. The compactness of classes can be calculated using different criteria, such as standard deviation, internal clustering indices or distance measures. To calculate compactness measure we created a weight function to be used as cost function for evaluating the goodness of every classifier, then returned the best classifier as a final result for the optimization process. After this process the items can be classified by the best rule-based classifier values.

For a generalised optimisation process, assume that experts' definitions and consultation produce **N** classes for items that have to be classified using aggregated attributes of temporal data, producing **D** of possible classifiers of rule-based classification for different ranges of values for each class. Our task is to select the best classifier among a set S of size **D** classifiers, hence reducing each provided separator range between neighbouring classes into a single line of separation, using the temporal attributes of T time points. A cost function $f(C)$ for each C ∈ S can be produced using any compact measure (CM) that measures the goodness of classes in each time point. The $f(C)$ can be defined as:

$$f(C) = \sum_{t=1}^{T} \sum_{n=1}^{N} CM(c_n^t) \times |c_n|$$

Where $|c_n|$ is number of items in each class to prevent creating single big classes. The classifier with the smallest $f(C)$ value can be considered as the best classifier among S.

$$BestClassifier = [C \text{ for any } C \in S \text{ with}$$
$$f(C) = Min(\{f(S[i])\}) Where\ i = 1\ to\ D]$$

As a result, in our example for students' classification according to their grades, the generated rules will be optimised by applying the proposed method of using temporal data to evaluate the goodness of the classes. The better agglomeration of class member in each time point, the better the classification is. As the data is univariate, standard deviation can be used as a CM measure, so that after the end of the optimisation process we might have rules like:

- If average(mark) ≥ 76 then: Excellent student
- If average(mark) ≤ 52 then: Bad student
- Else: Good Student

### 5. CASE STUDY

As mentioned previously, the our case study date originated from an experiment conducted by Fischbacher et al. (Fischbacher et al. 2012). The experiment was presented as a game of public good which can be played by four participants, each starting with 20 tokens representing real money. During the game players can keep these tokens or contribute with any amount of it in a project which represents a public good. After all players make their decision for the contribution, one round of the game is finished and players' gain will be uncovered for them. The amount of individual gain equals the amount of tokens kept by the player plus a 0.4 of the amount of project:

$$payoff_i = 20 - g_i + 0.4 \sum_{j=1}^{4} g_j$$

Where $payoff_i$ is player's gain from the round, $g_i$ is player's own contribution and $g_j$ is each player's contribution.

*A. Collected Data*

The collected data consists of 140 players, each of whom played for 10 rounds. The attributes of the data are listed below. Note that the player belief, contribution and other players' contributions are the only temporal attributes:

**Player type**: using Fischbacher et al.'s classification (Fischbacher et al. 2001). Player

types, average of contribution and percentage of each type are shown in Figure 3.

**Player identifier**: unique identifier though all sessions.

**Game controls**: attributes of session number, period of the game and game sequence.

**Belief**: player's belief about other players' rounded average contribution.

**Contribution**: Player's own contribution to the project in the current round.

**Other players' contribution**: Other players' rounded average of contribution.

**Contribution table**: 21 attributes represent initial willingness to contribute in regard to other players' contributions to public good projects. These attributes are filled once by players asked to state their hypothetical contribution, assuming a rounded average of other co-players contribution.

**Predicted contribution**: supposed contribution according to players' belief and contribution table.

The original classes for players created by economists are purely based on players' contribution preferences, regardless of their behaviour in the game, as they are classified according to the contribution table, which means the temporal data is not used in the classification. In the next section we will try to use the temporal dimension to classify players in a different perspective.

*B. Classifying Players Using Proposed Method*

To use the proposed method for classification we started with the economist's classification (Fischbacher et al. 2001), as they are experts in this field. With the help of the authors of the original classes for the public good game we presented definitions of the classes to fit with the temporal nature of our classification method and to be specific for the underlying data which has to be classified. The new definitions and classes are:

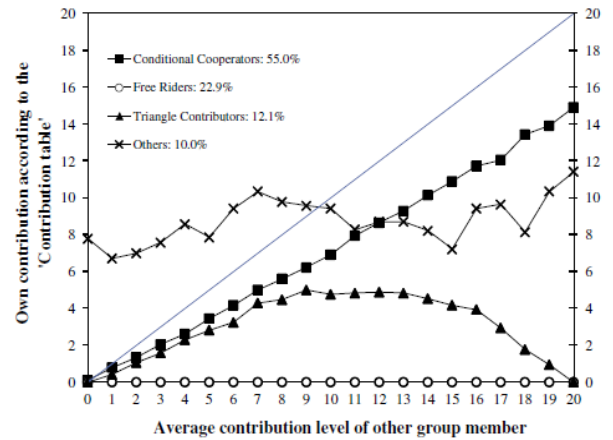

Figure 3. Four types of players' average own contribution according to co-players average contribution as labelled by economists.

**Free Riders**: players who contribute by equal or less than one point on average for all rounds or who are not contributing in most rounds. This class corresponds to the traditional category of Free Riders.

**Weak Contributors**: players who contribute between 1 and 5 or they are not contributing in half of the rounds. In the old categorisation, this class loosely relates to conditional contributors.

**Normal Contributors**: players who contribute on average around 5 points. This class is strongly related to conditional contributors as it fits the same criteria.

**Strong Contributors**: players who contribute more than 10 points on average. This class relates to conditional comparators and others in the classical categories.

To create the classifier rules based on these definitions, we generated the following aggregated attributes derived from the temporal attributes of the data:

**Contribution Mean**: players' contribution average for all rounds.

**Belief Mean**: players' average belief about co-players' contributions for all rounds.

**Number of Zero Contribution**: Number of game rounds in which the player contributed by zero points to the public good project.

The values for ranges which are presented to the rule-based classification optimizer are shown in Table 1. The lower and upper values represent the ranges and the best is the result value obtained from optimiser. We used brute forcing to find the best value. However it is possible to use a heuristic method like Differential Evolution (Storn & Price 1947). Figure 4 shows the resultant classes.

Table 1 List of provided upper and lower values as well as best results of the ranges

|       | Average Contribution | | | Average Belief | | | Zero Contribution | |
|-------|----|----|----|----|----|----|----|----|
|       | *Fr* | *Wc* | *Nc* | *Fr* | *Wc* | *Nc* | *Fr* | *Wc* |
| **Lower** | 0 | 1 | 2 | 2 | 4 | 2 | 6 | 5 |
| **Upper** | 1 | 4 | 6 | 9 | 9 | 9 | 9 | 7 |
| **Best**  | 1 | 1 | 6 | 2 | 5 | 6 | 9 | 6 |

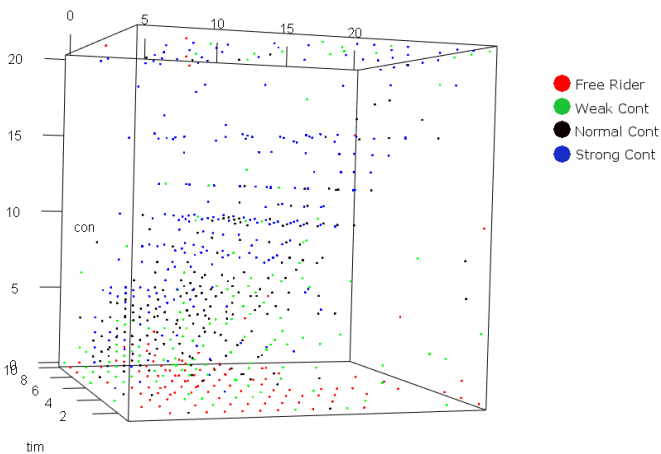

Figure 4. A 3D representation of players' contributions and beliefs over 10 rounds with different colours representing their classes

## 6. COMPARING RESULTS

To assess the generated results using the proposed method we conducted multiple tests and comparisons. We started by comparing original classes with the generated ones by plotting players' contributions for each category, then we presented the percentage of agreement between two classifications, and finally we examined the accuracy and how much they reflect the actual players' behaviour by testing them against classification and clustering algorithms and the amount of accurate predictability by which these algorithms correctly categorised the players, which demonstrates the accuracy of the category to group players correctly.

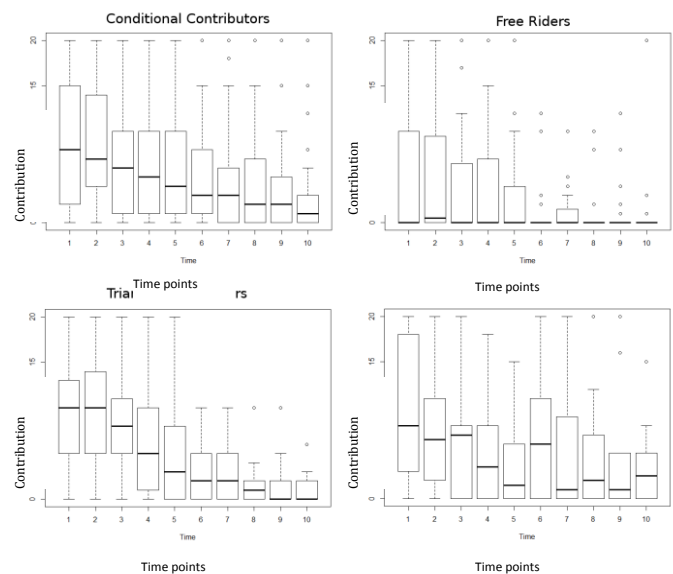

Figure 5. Means and box-plots of the players' contributions over ten rounds as generated using original economists' classification method

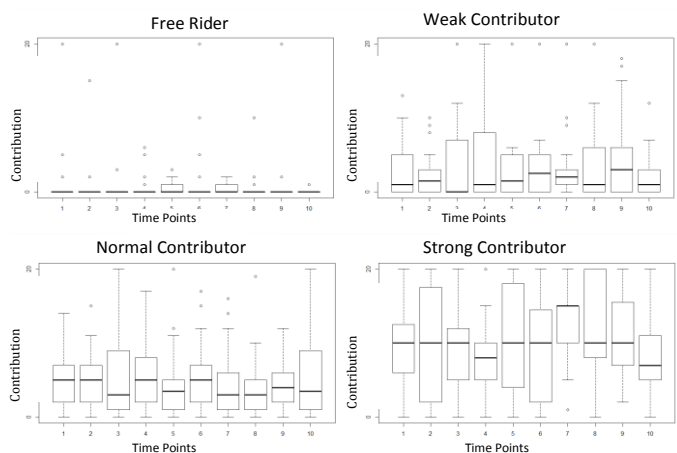

Figure 6 Means and box-plots of the players' contributions over ten rounds as generated proposed classification method.

To examine player behaviour though time points and present how closely they matched their classes plots for players' contribution of each classes, Figure 5 shows the original classification and Figure 6 shows the new categories. It can be seen that there is a stronger correlation between the new class labels and players' amount of contribution than the original labels.

Another test to compare categories is agreement of players' labels (between any two of them) by counting the percentage of players' occurrence. Table 2 illustrates correlations between any two categories.

Table 2 Player's agreement (by percentage) on the related category of the original and new class labels

|  |  | New Classes | | | |
|---|---|---|---|---|---|
| **Original Classes** |  | *FR* | *WC* | *NC* | *SC* |
|  | *FR* | 56.2 | 21.9 | 15.6 | 6.2 |
|  | *CC* | 16.9 | 14.3 | 36.4 | 32.5 |
|  | *TC* | 17.6 | 23.5 | 35.3 | 23.5 |
|  | *O* | 21.4 | 28.6 | 21.4 | 28.6 |

The final test is to examine the accuracy of categories using another (third) classification test. We use SVM classification to test its ability to predict 25% of the players' labels as a test set by using remaining 75% players as training set. We assume that higher accuracy of detecting players' labels indicates that the provided labels for the players are more representative and consistent though time points. We used AUC to measure the accuracy of SVM. Moreover, new attributes derived from original attributes as well as the original ones are used for classification, to make sure that the SVM classifier has no bias in any set of attributes. The new derived attributes are:

**Payoff**: the amount of points which the individual player gets during each round. This can be calculated by points kept plus public goods project gain. This attribute might have a great impact on the next game rounds' strategy, and it may reveal the player's category.

**Initial Deviation**: the difference between actual contribution and supposed contribution regarding the player's belief. The importance of this attribute is to show how much players stay on the initial plan that they had.

**Initial Deviation Mean**: this attribute validates of player's initial claim of willingness for contribution.

Prediction Accuracy: the difference between player's belief about other players' contributions and their actual contributions. Hence belief – others Contribution

**Prediction Accuracy SD**: Standard deviation of prediction Accuracy for each player in 10 rounds. This attribute is important to see how much player knows or anticipate correctly about other co-players.

The classification using SVM is carried out on all ten time points using ten folds of cross validation. This means each value of AUC is an average of 100 results. Moreover, four different sets of attributes were used to classify each category separately. The first set contains only players' beliefs and contribution attributes, which exist in the original dataset. The second set is the original set of data as collected by economists during the public goods game experiment, excluding control attributes (as explained in 5.1). The third set of attributes is the new derived attributes plus average of contribution and belief of each player, as described above. The last set is all attributes original and derived ones.

The results in Table 3 show that the proposed categories are better detected by the SVM classifier and there is a significant difference between the original and new labels, especially using only belief and contribution attributes. However, the results for other attribute sets are not as significant as the former one.

These results show that the new proposed method might be beneficial for providing categories for temporal data reflecting players' behaviour during different time points. The results of belief and contribution show that the original classifier cannot perform very well, as these two attributes are mainly temporal, and the original categories are not designed to function with temporal data.

Table 3 Classification performance (mean of AUC) for different attribute sets

| Attributes | Original Classes | New Classes |
|---|---|---|
| *Belief & Contrib* | 0.502 | 0.748 |
| *Original* | 0.74 | 0.764 |
| *Derived* | 0.662 | 0.827 |
| *Original & Derived* | 0.741 | 0.801 |

For the original dataset the results shows a significant improvement of the original categories results, as the data contains a contribution table which is the basis of its classification, while it does not improve the proposed classification with the same rate as it does not take advantage of the table.

The results of the derived attributes show better performance for the new classifier and slightly worse result for the original classifier labels due to the nature of the derived dataset, as most of the attributes used temporal attributes as the source for their data.

The final row of results, which are the combination of both original and derived datasets, show no significant change of the original classes compared with the original dataset alone, while the new classifier has degraded results compared with the derived attributes. This again confirms that the contribution table does not increase the performance of the new classifier, while the temporal attributes cannot be used as a valid attributes with the original classifier. Categorizing players according to their behaviour might open a new opportunity to study changes in their behaviour during a series of game play.

**Acknowledgements**

The authors record their thanks to Simon Gaechter and Felix Kolle in the School of Economics at the University of Nottingham for providing us with data from the public goods experiment and taking time to explain it to us.